\documentclass{article}

    \PassOptionsToPackage{numbers, compress}{natbib}

     \usepackage[final]{neurips_2020}

\usepackage[utf8]{inputenc} 
\usepackage[T1]{fontenc}    
\usepackage{hyperref}       
\usepackage{url}            
\usepackage{booktabs}       
\usepackage{amsfonts}       
\usepackage{amssymb}
\usepackage{nicefrac}       
\usepackage{microtype}      
\usepackage{graphicx}
\usepackage{relsize}
\usepackage{amsmath}
\usepackage{subfigure}
\usepackage{esvect}

\title{Network Bending: Expressive Manipulation of Deep Generative Models}

\author{%
  Terence Broad \\
  Department of Computing\\
  Goldsmiths, University of London\\
  \texttt{t.broad@gold.ac.uk} \\
  \And
   Frederic Fol Leymarie \\
  Department of Computing\\
  Goldsmiths, University of London\\
  \texttt{ffl@gold.ac.uk} \\
   \AND
   Mick Grierson \\
  Creative Computing Institute\\
  University of The Arts London\\
  \texttt{m.grierson@arts.ac.uk} \\
}

\begin{document}

\maketitle

\begin{abstract}
We introduce a new framework for manipulating and interacting with deep generative models that we call \textit{network bending}. We present a comprehensive set of deterministic transformations that can be inserted as distinct layers into the computational graph of a trained generative neural network and applied during inference. In addition, we present a novel algorithm for analysing the deep generative model and clustering features based on their spatial activation maps. This allows features to be grouped together based on spatial similarity in an unsupervised fashion. This results in the meaningful manipulation of sets of features that correspond to the generation of a broad array of semantically significant features of the generated images. We outline this framework, demonstrating our results on state-of-the-art deep generative models trained on several image datasets. We show how it allows for the direct manipulation of semantically meaningful aspects of the generative process as well as allowing for a broad range of expressive outcomes.

\end{abstract}

\section{Introduction}

We introduce a new framework for the direct manipulation of deep generative models that we call \textit{network bending}. This framework allows for \textit{active divergence} \cite{berns2020bridging} from the original training distribution in a flexible way that provides a broad range of expressive outcomes. We have implemented a wide array of image filters that can be inserted into the network and applied to any assortment of features, in any layer, in any order.  We use a plug-in architecture to dynamically insert these filters as individual layers inside the computational graph of the pre-trained generative neural network, ensuring efficiency and minimal dependencies. As this process is altering the computation graph of the model, changes get applied to the entire distribution of generated results. We also present a novel approach to grouping together features in each layer. This is based on the spatial similarity of the activation map of the features and is done to reduce the dimensionality of the parameters that need to be configured by the user. It gives insight into how \textit{groups} of features combine to produced different aspects of the image. We show results from these processes performed on generative models for images (using StyleGAN2, the current state-of-the-art for unconditional image generation \cite{karras2019analyzing}) trained on several different datasets, and map out a pipeline to harness the generative capacity of deep generative models in producing novel and expressive outcomes.

\section{Related Work}

\subsection{Deep Generative Models}

A generative model consists in the application of machine learning to learn a configuration of parameters that can approximately model a given data distribution. This was historically a very difficult problem, especially for domains of high data dimensionality such as for audio and images. With the advent of deep learning and large training datasets, great advances were made in the last decade. Deep neural networks are now capable of generating realistic audio \cite{oord2016wavenet,dhariwal2020jukebox} and images \cite{brock2018large,karras2019style,karras2019analyzing}. In the case of images, Variational Autoencoders \cite{kingma2013auto,rezende2014stochastic} and Generative Adversarial Networks (GANs) \cite{goodfellow2014generative} have been major breakthroughs that provide powerful training methods. Over the past few years there has been major improvements to their fidelity and training stability, with application of convolutional architecture \cite{radford2015unsupervised}, progressively growing architecture \cite{karras2017progressive}, leading to the current state of the art in producing unconditional photo-realistic samples in StyleGAN \cite{karras2019style} and then StyleGAN2 \cite{karras2019analyzing}. One class of conditional generative models that take inputs in the form of semantic segmentation maps can be used to perform semantic image synthesis, where an input mask is used to generate an image of photographic quality \cite{isola2017image, chen2017photographic, park2019semantic}.

Understanding and manipulating the \emph{latent space} of generative models has subsequently been a growing area of research. Semantic latent manipulation consists in making informed alterations to the latent code that correspond to the manipulation of different semantic properties present in the data. This can be done by operating directly on the latent codes \cite{brock2016neural, shen2020interpreting} or by analysing the activation space of latent codes to discover interpretable directions of manipulation in latent space \cite{harkonen2020ganspace}. Evolutionary methods have been applied to search and map the latent space \cite{bontrager2018deepmasterprints, fernandes2020evolutionary} and interactive evolutionary interfaces have also been built to operate on the latent codes \cite{Simon-ganbreeder} for human users to explore and generate samples from generative models.

\subsection{Analysis of Deep Neural Networks}

Developing methods for understanding the purpose of the internal features (aka hidden units) of deep neural networks has been an on-going area of research. In computer vision and image processing applications, there have been a number of approaches, such as through visualisation, either by sampling patches that maximise the activation of hidden units \cite{zeiler2014visualizing, zhou2014object}, or by using variations of backpropagation to generate salient image features \cite{zeiler2014visualizing, simonyan2013deep}. A more sophisticated approach is \textit{network dissection} \cite{Bau2018-td} where hidden units responsible for the detection of semantic properties are identified by analysing their responses to semantic concepts and quantifying their alignment. Network dissection was later adapted and applied to generative models \cite{Bau2018-td}, by removing individual units, while using in combination a bounding box detector trained on the ADE20K Scene dataset \cite{zhou2017scene}. This led to the ability to identify a number of units associated with the generating of certain aspects of the scene. This approach has since been adapted for music generation \cite{Brink2019-gc}. 

\subsection{Manipulation of Deep Generative Models}

The manipulation of deep generative models is itself a nascent area of research. An interactive interface built upon the GAN Dissection approach  \cite{Bau2018-td} was presented with the GANPaint framework in 2019 \cite{bau2019semantic}. This allows users to `paint' onto an input image in order to edit and control the spatial formation of hand-picked features generated by the GAN. 

An approach that alters the computational graph of the model such that a change alters the entire distribution of results, is presented as an algorithm for
``rewriting the rules of a generative model'' \cite{bau2020rewriting}. In this approach, the weights from a single convolutional layer are used as an associative memory. Using a copy-paste interface, a user can then map a new element onto a generated output. The algorithm uses a process of constrained optimisation to edit values in the weight matrix to find the closest match to the copy-paste target. Once the rules of the weight matrix have been altered, all results from the generator have also been altered. 




\section{Clustering Features}
\label{section:clustering}

As most of the layers in current state of the art GANs, such as StyleGAN2, have very large numbers of convolutional features, controlling each one individually would be far too complicated to build a user interface around and to control these in a meaningful way. In addition, because of the redundancy existing in these models, manipulating individual features does not normally produce any kind of meaningful outcome. Therefore, it is necessary to find some way of grouping them together into more manageable ensembles of sets of features. Ideally such sets of features would correspond to the generation of distinct, semantically meaningful aspects of the image, and manipulating each set would correspond to the manipulation of specific semantic properties in the resulting generated sample. In order to achieve this, we present a novel approach, combining metric learning and a clustering algorithm to group sets of features in each layer based on the spatial similarity of their activation maps. We train a separate convolutional neural network (CNN) for each layer of the StyleGAN2 model with a bottleneck architecture (first introduced by Gr{\'e}zl et al.~\cite{grezl2007probabilistic}) to learn a highly compressed feature representation; the later is then used in a metric learning approach in combination with the $k$-means clustering algorithm \cite{lloyd1982least, celebi2013comparative} to group sets of features in an unsupervised fashion. In our experiments we have performed this feature clustering process on models trained on three different datasets: the FFHQ \cite{karras2019style}, LSUN churches and LSUN cats datasets \cite{yu2015lsun}.

\subsection{Architecture}

For each layer of the StyleGAN2 model, we train a separate CNN on the activation maps of all the convolutional features. As the resolution of the activation maps and number of features varies for the different layers of the model (a breakdown of which can be seen in Table \ref{tab:classifier-table}) we employ an architecture that can dynamically be changed, by increasing the number of convolutional blocks, depending on what depth is required. 

\begin{table*}[]
\centering
\begin{tabular}{|c|c|c|c|c|c|}
\hline
Layer & Resolution & \#features &  CNN depth & \#clusters & Batch size\\
\hline
1     & 8x8        & 512          & 1                & 5                & 500        \\
2     & 8x8        & 512          & 1                & 5                & 500        \\
3     & 16x16      & 512          & 2                & 5                & 500        \\
4     & 16x16      & 512          & 2                & 5                & 500        \\
5     & 32x32      & 512          & 3                & 5                & 500        \\
6     & 32x32      & 512          & 3                & 5                & 500        \\
7     & 64x64      & 512          & 4                & 5                & 200        \\
8     & 64x64      & 512          & 4                & 5                & 200         \\
9     & 128x128    & 256          & 5                & 4                & 80         \\
10    & 128x128    & 256          & 5                & 4                & 80         \\
11    & 256x256    & 128          & 6                & 4                & 50         \\
12    & 256x256    & 128          & 6                & 4                & 50         \\
13    & 512x512    & 64           & 7                & 3                & 20         \\
14    & 512x512    & 64           & 7                & 3                & 20         \\
15    & 1024x1024  & 32           & 8                & 3                & 10         \\
16    & 1024x1024  & 32           & 8                & 3                & 10       \\
\hline
\end{tabular}
\medskip
\caption{\label{tab:classifier-table}Table showing resolution, number of features of each layer, the number of ShuffleNet \cite{zhang2018shufflenet} convolutional blocks for each CNN model used for metric learning, the number of clusters calculated for each layer using $k$-means and the batch size used for training the CNN classifiers. Note: LSUN church and cat models have only 12 layers.
}
\end{table*}

We employ the ShuffleNet architecture \cite{zhang2018shufflenet} for the convolutional blocks in the network, which is one of the state-of-the-art architectures for efficient inference (in terms of memory and speed) in computer vision applications. For each convolutional block we utilise a feature depth of 50 and have one residual block per layer. The motivating factor in many of the decisions made for the architecture design was not focused on achieving the best accuracy per se. Instead, we want a network that can learn a sufficiently good metric while also being reasonably quick to train (with 12-16 separate classifiers per GAN model). We also want a lightweight enough network, such that it could be used in a real-time setting where clusters can quickly be calculated for an individual latent encoding, or it could be used efficiently when processing large batches of samples.

After the convolutional blocks, we flatten the final layer (4x4x50) and learn from it a mapping into a narrow bottleneck ($\vv{v} \in \mathbb{R}^{10}$), before re-expanding the dimensionality of the final layer to the number of convolutional features present in the GAN layer. The goal of this bottleneck is to force the network to learn a highly compressed representation of the different convolutional features in the GAN. While this invariably looses some information, most likely negatively affecting classification performance during training, this is in-fact the desired result. We want to force the CNN to combine features of the activation maps with similar spatial characteristics so that they can easily be grouped together by the clustering algorithm. Another motivating factor is that the clustering algorithm we have chosen ($k$-means) does not scale well for feature spaces with high dimensionality.

\subsection{Training}

We generated a training set of the activations of every feature for every layer of 1000 randomly sampled images, and a test set of 100 samples for the models trained on all of the datasets used in our experiments. We trained each CNN using the softmax feature learning approach \cite{dosovitskiy2014discriminative}, a reliable method for distance metric learning. This method employs the standard softmax training regime \cite{bridle1990probabilistic} for CNN classifiers. Each classifier has been initialised with random weights and then trained for 100 epochs using the Adam optimiser \cite{kingma2014adam} with a learning rate of 0.0001 and with $\beta_1 = 0.9$ and $\beta_2 = 0.999$. All experiments were carried out on a single NVIDIA GTX 1080ti. The batch size used for training the classifiers for the various layers can be seen in Table \ref{tab:classifier-table}.

After training, the softmax layer is discarded and the embedding of the final layer is used as the discriminative feature vector where the distances between points in feature space permit to gauge the degree of similarity of two samples. The one difference in our approach to standard softmax feature learning is that we use the second to last layer, the feature vector from the bottleneck, giving a more compressed feature representation than what standard softmax feature learning would offer.

\subsection{Clustering Algorithm}

Once the CNNs for every layers have been trained, they can then be used to extract feature representations of the activation maps of the different convolutional features corresponding to each individual layer of the GAN. There are two approaches to this. The first is to perform the clustering on-the-fly for a specific latent for one sample. A user would want to do this to get customised control of a specific sample, such as a latent that has been found to produce the closest possible reproduction of a specific person from the StyleGAN2 model trained on the FFHQ dataset \cite{abdal2019image2stylegan,karras2019analyzing}. The second approach is to perform clustering based on an average of features' embedding drawn from many random samples, which can be used to find a general purpose set of clusters.

The clustering algorithm for a single example is activated by a forward pass of the GAN performed without any additional transformation layers being inserted, this to obtain the unmodified activation maps. The activation map $X_{df}$ for each layer $d$ and feature $f$ is fed into the CNN metric learning model for that layer $C_d$ to get the feature vector $\vv{v}_{df}$. The feature vectors for each layer are then aggregated and fed to the $k$-means clustering algorithm --- using Lloyd's method \cite{lloyd1982least} with Forgy initialization \cite{forgy1965cluster, celebi2013comparative}. This results in a pre-defined number of clusters for each layer. Sets of features for each layer can then be manipulated in tandem by the user.

Alternatively, to find a general purpose set of clusters, we first calculate the mean feature vector $\vv{\bar{v}}_{df}$ that describes the spatial activation map for each convolutional feature in each layer of StyleGAN2 from a set of $N$ randomly generated samples --- the results in the paper are from processing 1000 samples. Then we perform the same clustering algorithm as previously for individual samples on the mean feature vectors. 



\section{Transformation Layers}

We have implemented a broad variety of deterministically controlled transformation layers that can be dynamically inserted into the computational graph
of the generative model. The transformation layers are implemented natively in PyTorch \cite{paszke2019pytorch} for speed and efficiency. We treat the activation maps of each feature of the generative model as 1-channel images in the range -1 to 1. Each transformation is applied to the activation maps individually before they are passed to the next layer of the network. The transformation layers can be applied to all the features in a layer, or a random selection, or by using pre-defined groups automatically determined based on spatial similarity of the activation maps (Section \ref{section:clustering}). Figure \ref{fig:layerwide_comparison} shows a comparison of a selection of these transformations applied to all the features layer-wide in various layers.

\subsection{Numerical Transformations}

We begin with simple numerical transformations $f(x)$ that are applied to individual activation units $x$. We have implemented four distinct numerical transformations: the first is \emph{ablation}, which can be interpreted as $f(x) = x \cdot 0$. The second is \emph{inversion}, which is implemented as $f(x) = 1 - x$. The third is \emph{multiplication by a scalar} $p$ implemented as $f(x) = x \cdot p$. The final transformation is \emph{binary thresholding} (often referred to  as posterisation) with threshold $t$, such that:
\begin{equation}
f(x) = \begin{cases}
    1,& \text{if } x\geq t\\
    0,              & \text{otherwise}
\end{cases}
\end{equation}

\subsection{Affine Transformations}
\label{sec:affine}
For this set of transformations we treat each activation map $X$ for feature $f$ as an individual matrix, that simple affine transformations can be applied too. The first two are horizonal and vertical \emph{reflections} that are defined as:
\begin{equation}
X \begin{bmatrix}
-1 & 0 & 0\\
\ 0 & 1 & 0\\
\ 0 & 0 & 1
\end{bmatrix}\quad , \quad X \begin{bmatrix}
1 & \ 0 & 0\\
0 & -1 & 0\\
0 & \ 0 & 1
\end{bmatrix}
\end{equation}

\noindent The second is \emph{translations} by parameters $p_x$ and $p_y$ such that:
\begin{equation}
X \begin{bmatrix}
1 & 0 & p_x\\
0 & 1 & p_y\\
0 & 0 & 1
\end{bmatrix}
\end{equation}

\noindent The third is \emph{scaling} by parameters $k_x$ and $k_y$ such that:
\begin{equation}
X \begin{bmatrix}
k_x & 0 & 0\\
0 & k_y & 0\\
0 & 0 & 1
\end{bmatrix}
\end{equation}
Note that in this paper we only report on using uniform scalings, such that $k_x = k_y$. Finally, fourth is \emph{rotation} by an angle $\theta$ such that:
\begin{equation}
X \begin{bmatrix}
cos(\theta) & -sin(\theta) & 0\\
sin(\theta) & cos(\theta) & 0\\
0 & 0 & 1
\end{bmatrix}
\end{equation}

Other affine transformations can easily be implemented by designing the matrices accordingly.

\subsection{Morphological Transformations}

We have implemented two of the possible basic mathematical morphological transformation layers, performing \emph{erosion} and \emph{dilation} \cite{soille1999erosion} when applied to the activation maps, which can be interpreted as 1-channel images. These can be configured with the parameter $r$ which is the radius for a circular kernel (aka structural element) used in the morphological transformations.

\begin{figure*}[!ht]
    \centering
    \includegraphics[width=0.85\textwidth]{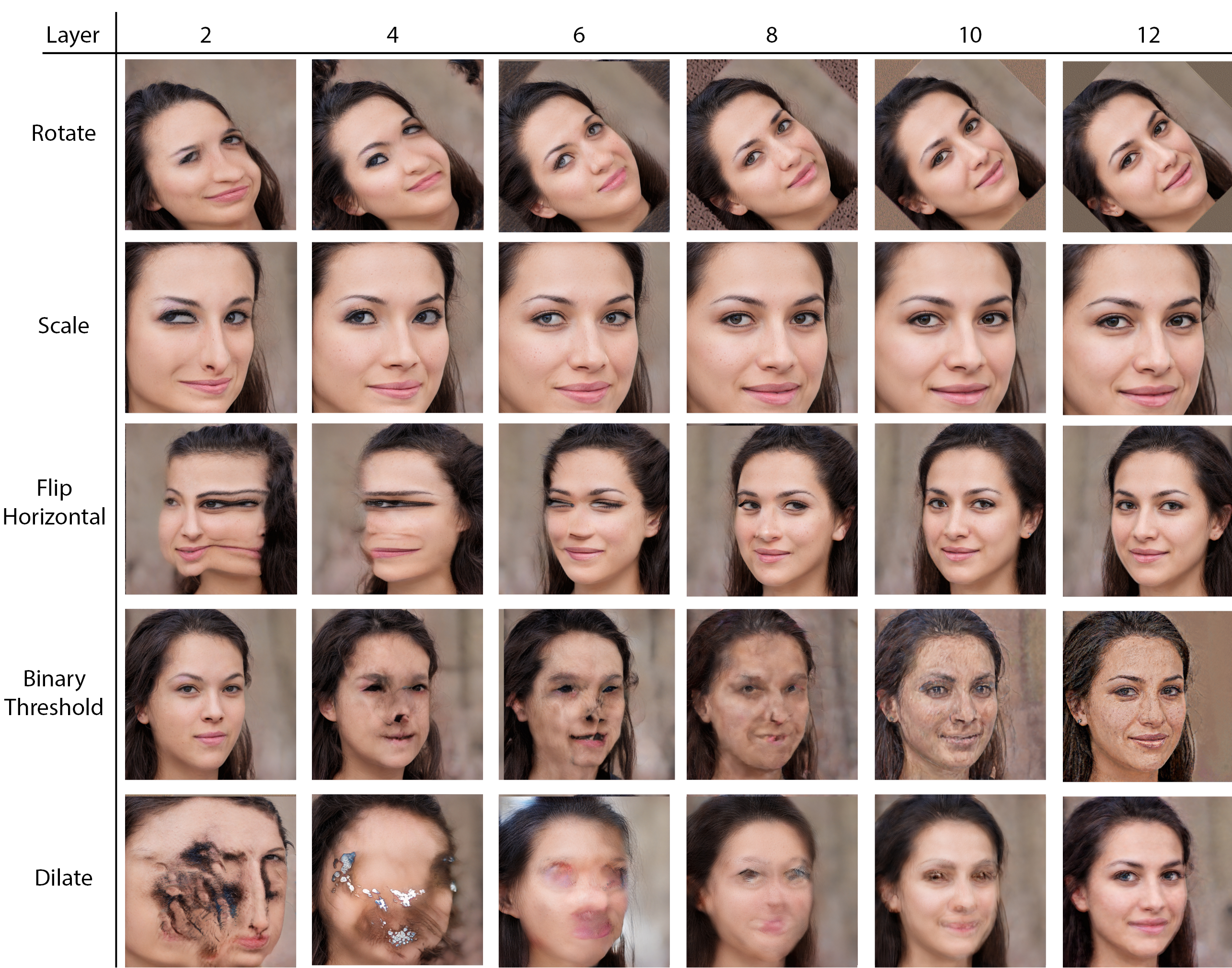}
    \caption{A comparison of various transformation layers inserted and applied to all of the features in different layers in the StyleGAN2 network trained on the FFHQ dataset, showing how applying the same filters in different layers can make wide-ranging changes the generated output. The rotation transformation is applied by an angle $\theta=45$. The scale transformation is applied by a factor of $k_{x}=k_{y}=0.6$. The binary threshold transformation is applied with a threshold of $t=0.5$. The dilation transformation is applied with a structuring element with radius $r=2$ pixels.}
    \label{fig:layerwide_comparison}
\end{figure*}

\section{Manipulation Pipeline}

In our current implementation,\footnote{Our implementation and the datasets we have used for training the clustering models are publicly available and can be found at: \url{https://github.com/terrybroad/network-bending}}\ transforms are specified in \texttt{YAML} configuration files \cite{ben2009yaml}, such that each transform is specified with 5 items: (i) the layer, (ii) the transform itself, (iii) the transform parameters, (iv) the layer type (i.e. how the features are selected in the layer: across all features in a layer, to pre-defined clusters, or to a random selection of features), and (v) the parameter associated with the layer type (either the cluster index, or the percentage of features the filter will randomly be applied to). There can be any number of transforms defined in such a configuration file.

\begin{figure*}[!ht]
    \centering
    \includegraphics[width=0.9\textwidth]{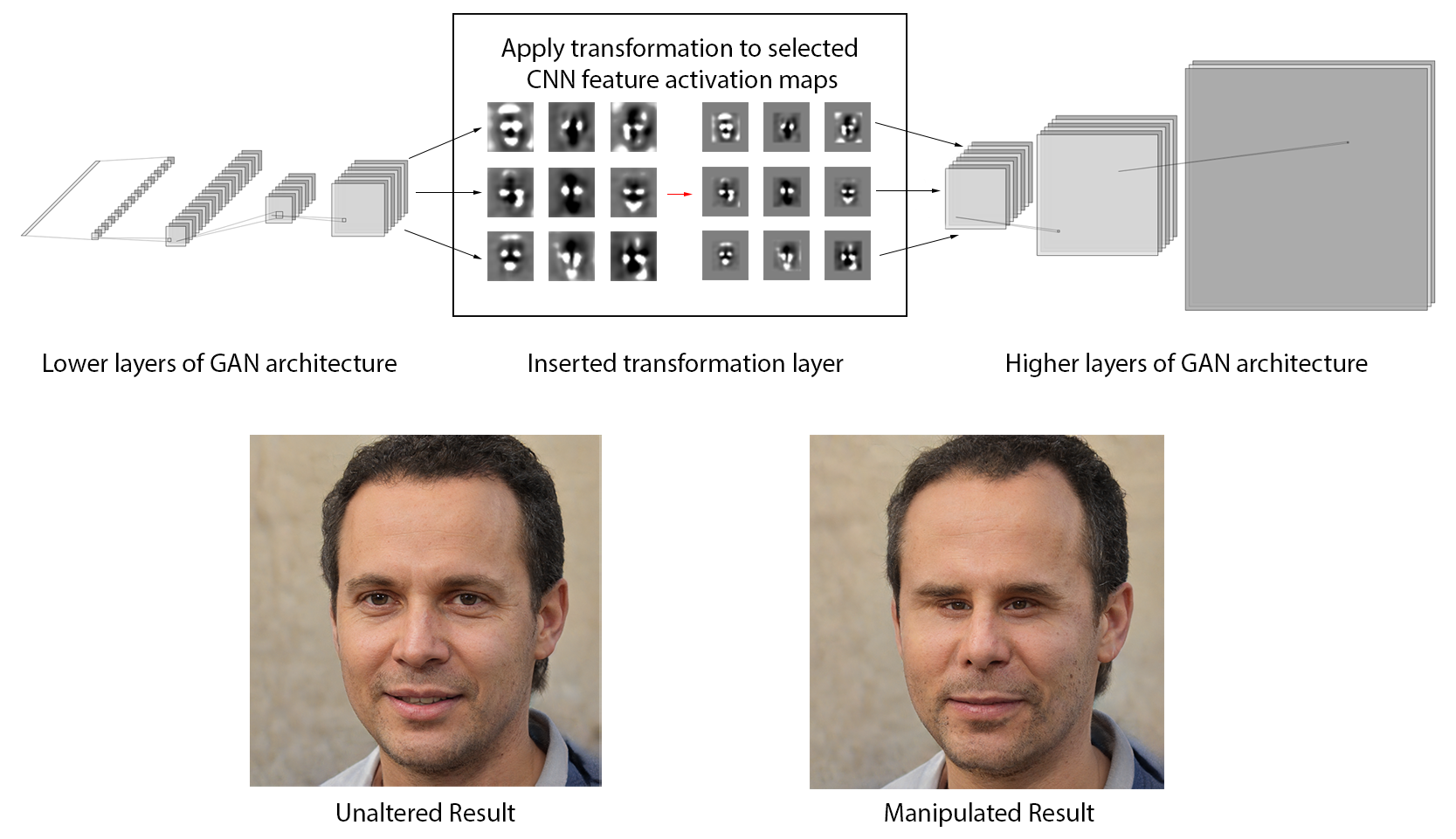}
    \caption{Overview of our \textit{network bending} approach where deterministically controlled transformation layers can be inserted into a pre-trained network. As an example, a transformation layer that scales the activation maps by a factor of $k_x=k_y=0.6$ is applied (\S\ref{sec:affine}) to a set of features in layer 5 responsible for the generation of eyes, which has been discovered in an unsupervised fashion using our algorithm to cluster features based on the spatial similarity of their activation maps (\S \ref{section:clustering}). At the bottom left we show the sample generated by StyleGAN2 \cite{karras2019analyzing} trained on the FFHQ dataset without modification, while to its right we show the same sample generated with the scaling transform applied to the selected features. NB: the GAN network architecture diagram shown on the top row is for illustrative purpose only.}
    \label{fig:overview_diagram}
\end{figure*}

After loading the configuration, we either lookup which features are in the cluster index, or randomly apply indices based on the random threshold parameter. Then the latent is loaded, which can either be randomly generated, or be predefined in latent space $z$, or be calculated using a projection in latent space $w$ \cite{abdal2019image2stylegan,karras2019analyzing}. The latent code is provided to the generator network and inference is performed. As our implementation is using PyTorch \cite{paszke2019pytorch}, a dynamic neural network library, these transformation layers can therefore be inserted dynamically during inference as and when they are required, and applied only to the specified features as defined by the configuration. Once inference in unrolled, the generated output is returned. Figure \ref{fig:overview_diagram} provides a visual overview of the pipeline, as well as a comparison between a modified and unmodified generated sample.

\subsection{Chaining Transformations}

\begin{figure}[!ht]
   \centering
   \subfigure{\includegraphics[width=.24\textwidth]{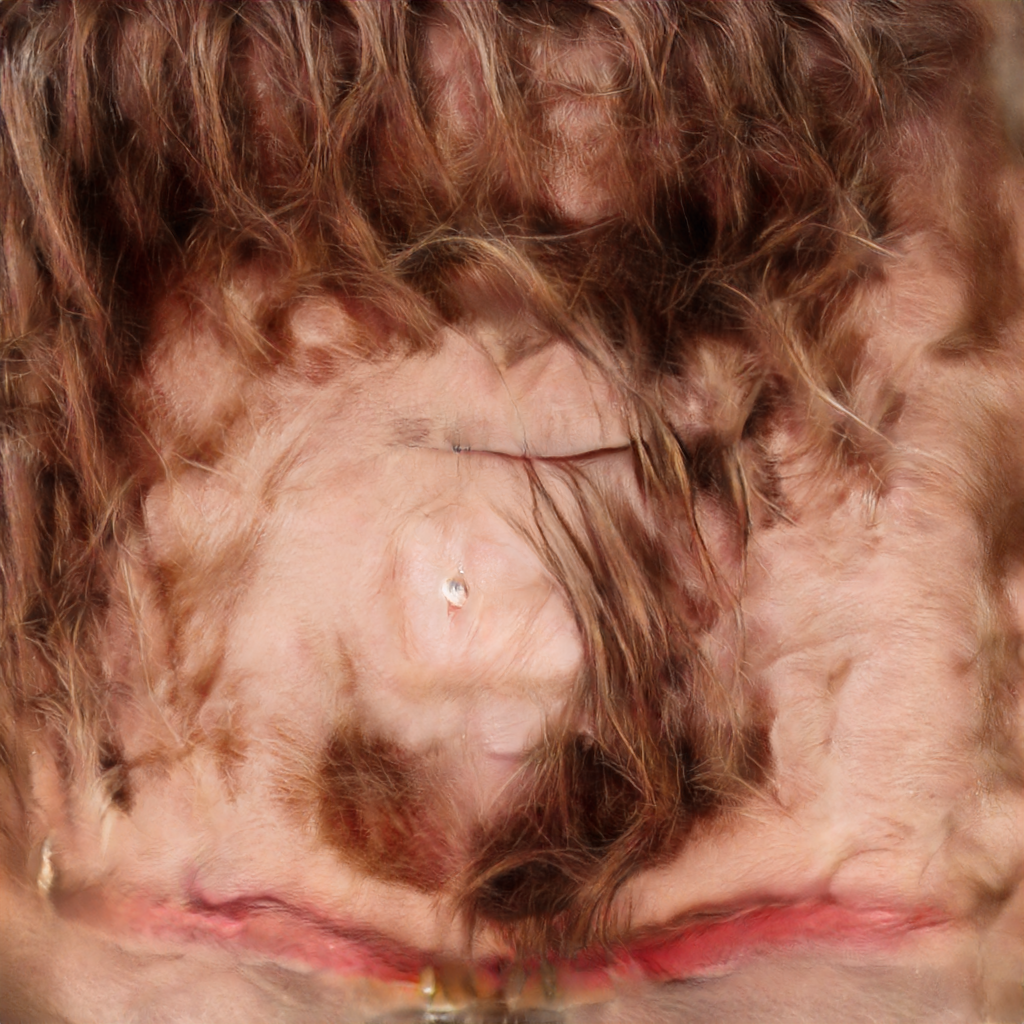}}
   \subfigure{\includegraphics[width=.24\textwidth]{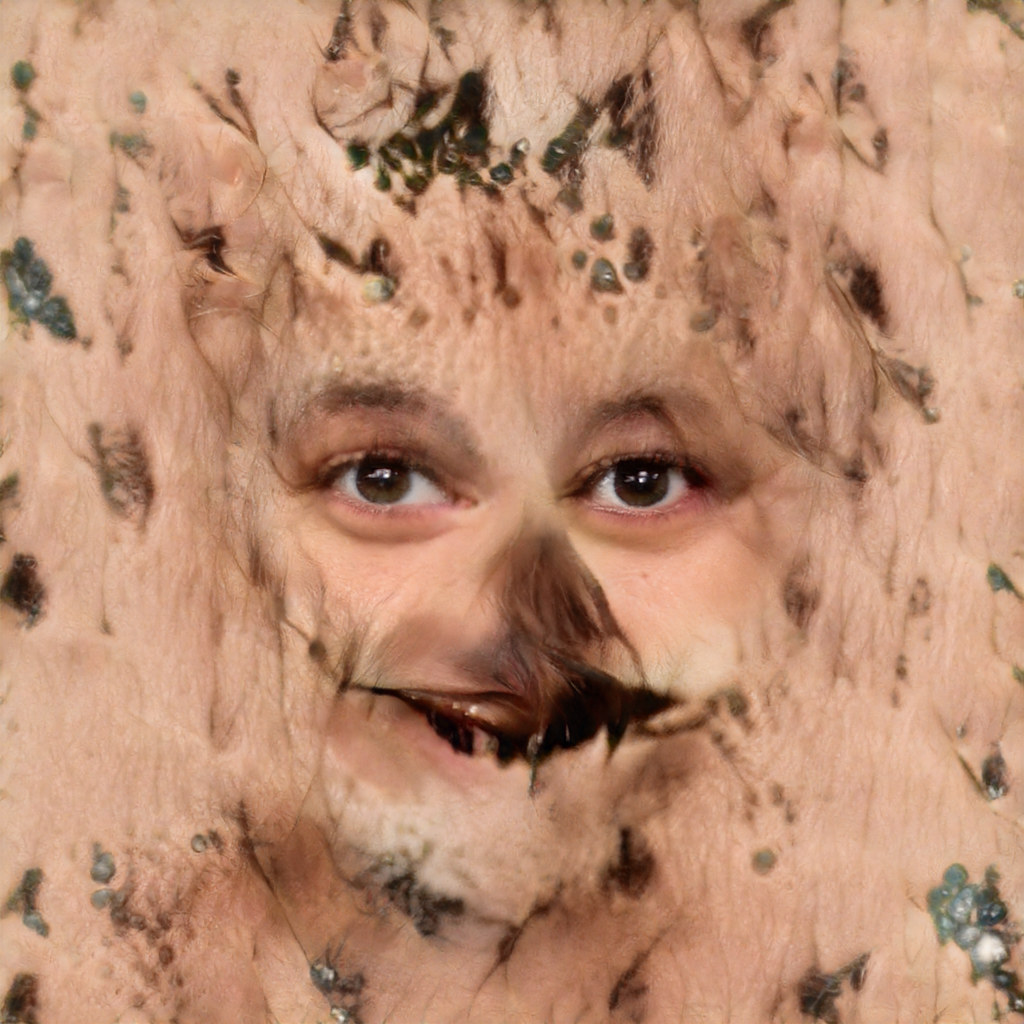}}
   \subfigure{\includegraphics[width=.24\textwidth]{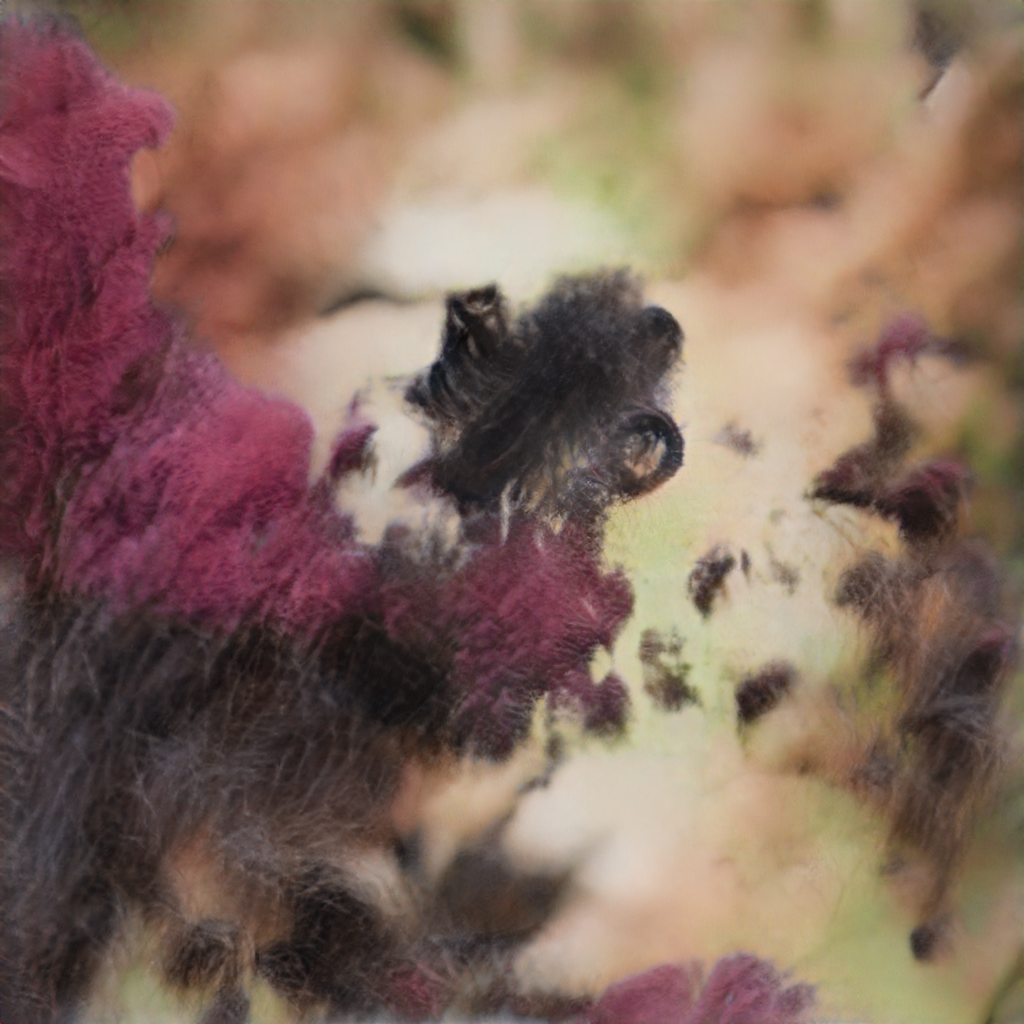}}
   \subfigure{\includegraphics[width=.24\textwidth]{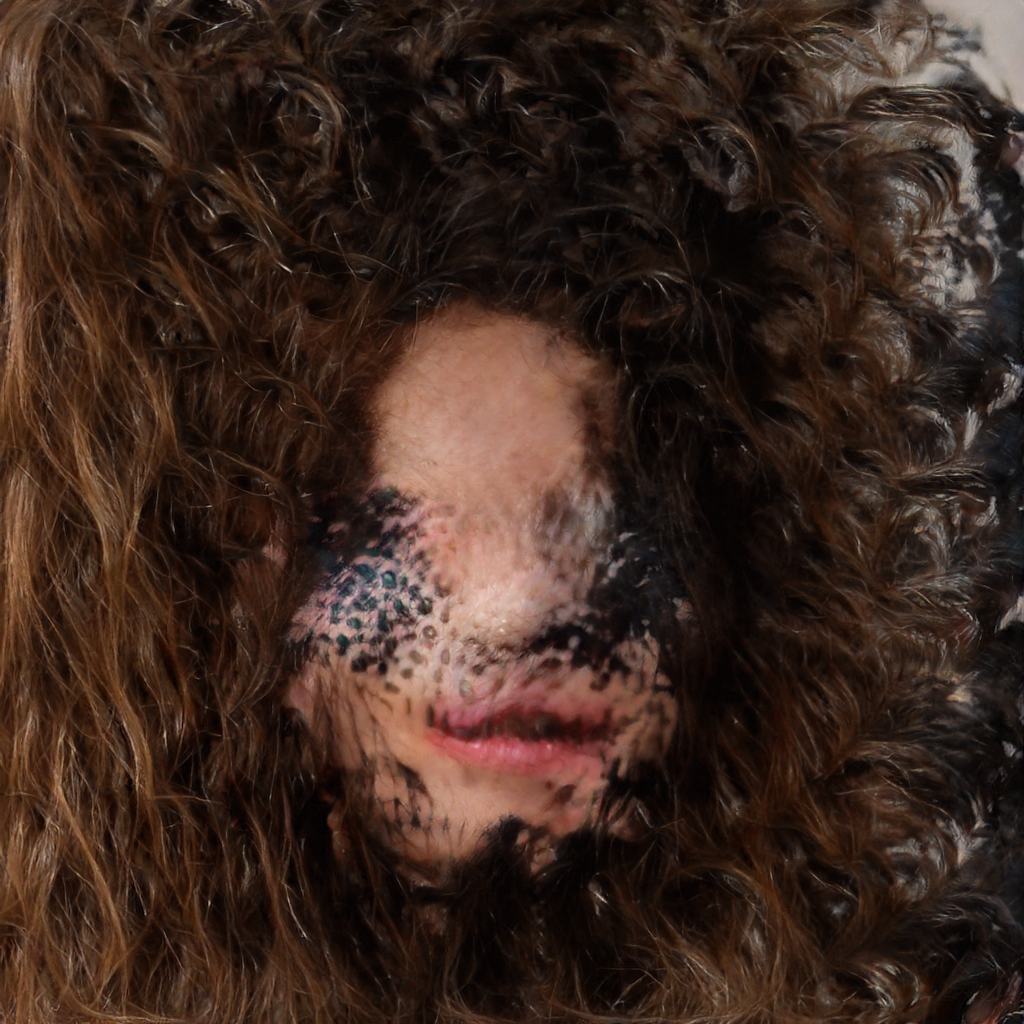}}
   \caption{A broad range of styles and novel outcomes can be achieved by chaining transformations. The 4 images show some samples of different configurations of transformations applied to different sets of features on different layers. All results are produced using the StyleGAN2 model trained on the FFHQ dataset. }
   \label{fig:chaining_transforms}
\end{figure}

From the perspective of building tools that impact the generation of expressive and novel samples, performing one transformation at a time can be quite restricting. With our approach, we are not limited in this manner, and a user can explore more complicated effects by chaining multiple transformations. In Figure~\ref{fig:chaining_transforms} a few examples of combining multiple transformations, when applied to different sets of features in different layers, illustrate how our proposed architecture can generate very unusual and highly distinctive results. This significantly broadens the space of possible outcomes to explore, allowing for surprising results when different transformations interact with each other. 

\subsection{Stochastic Layers}

Utilising stochastic layers, where the filters are applied to a random selections of features, can provide an alternative workflow than simply the straightforward direct manipulation of parameters for producing a single output. For instance, to take a real world example, these network bending techniques were used in the production of a series of EP (extended play record) artworks for the music band \emph{0171}, which provides an illustration of some of the affordances of the stochastic layers. A series of artworks were commissioned for 5 singles in an EP, such that the works had to be variations on the same theme, and while each artwork had to be unique, they shared a common visual aesthetic. A workflow was specifically developed to fit the brief and make use of possibilities afforded by the many variations on a theme made possible by using stochastic transformation layers. 

\begin{figure}[!ht]
   \centering
   \subfigure{\includegraphics[width=.25\textwidth]{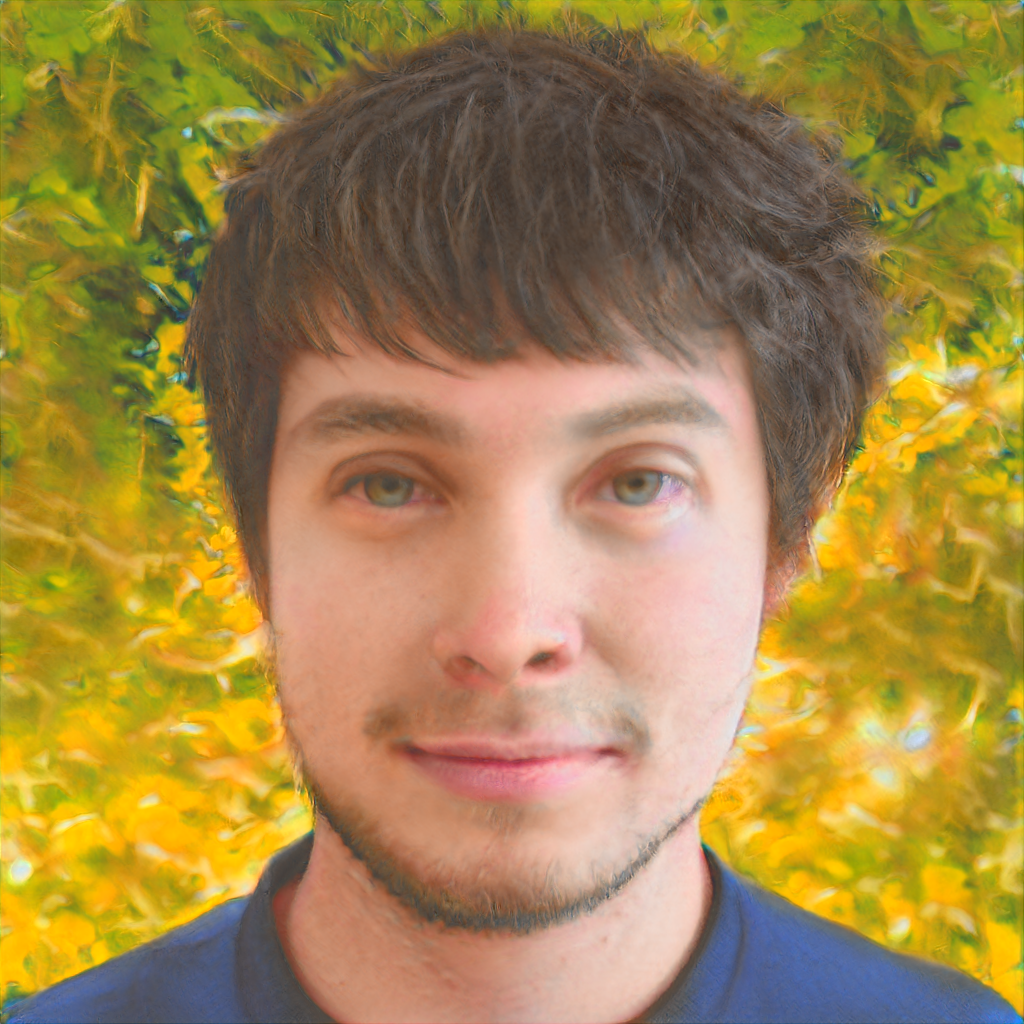}}
   \subfigure{\includegraphics[width=.25\textwidth]{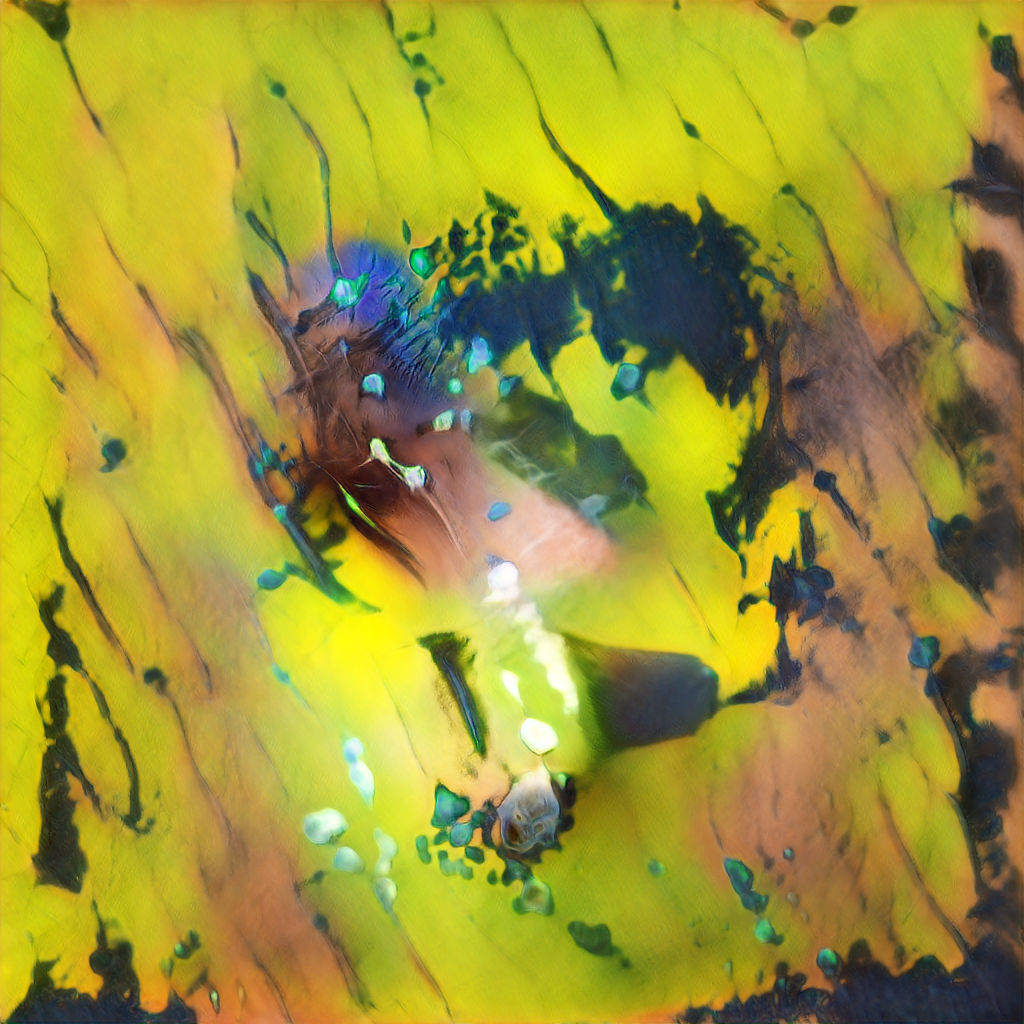}}
   \subfigure{\includegraphics[width=.25\textwidth]{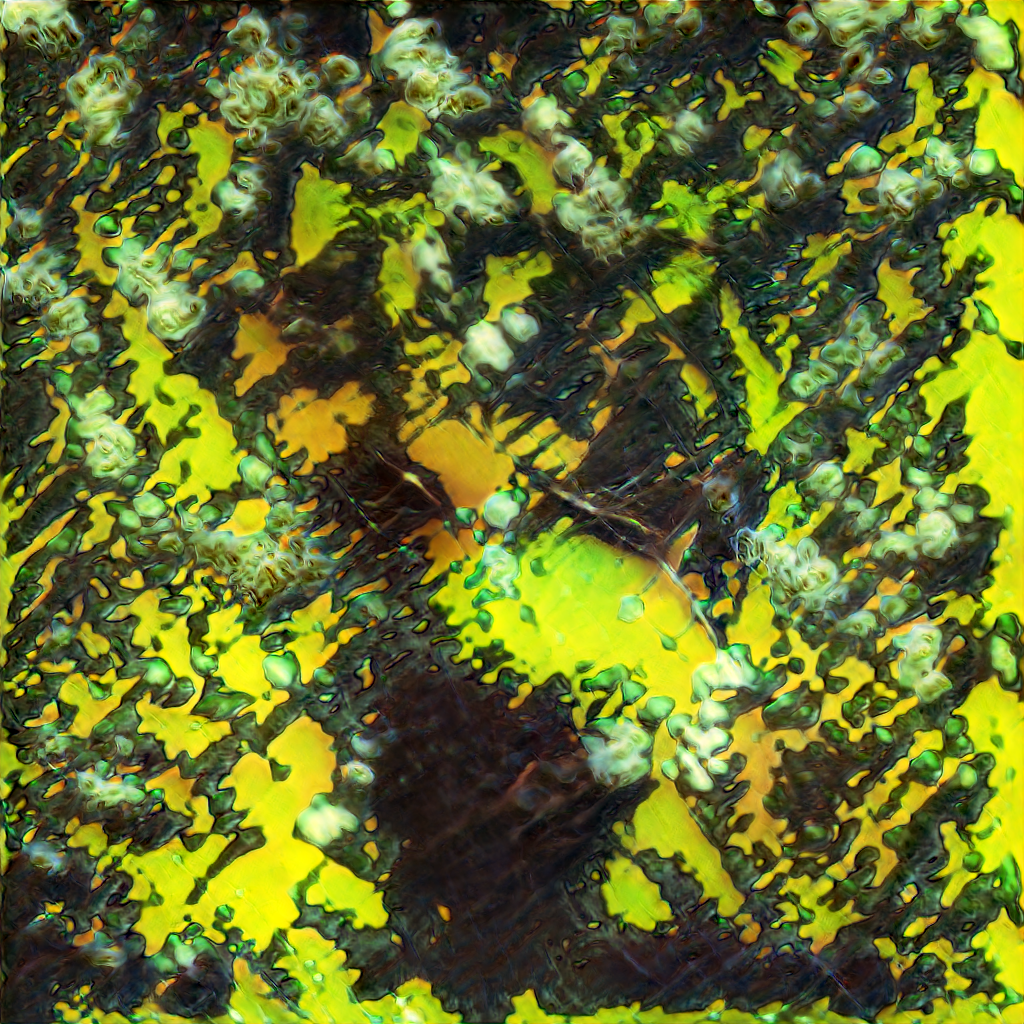}}
   \subfigure{\includegraphics[width=.25\textwidth]{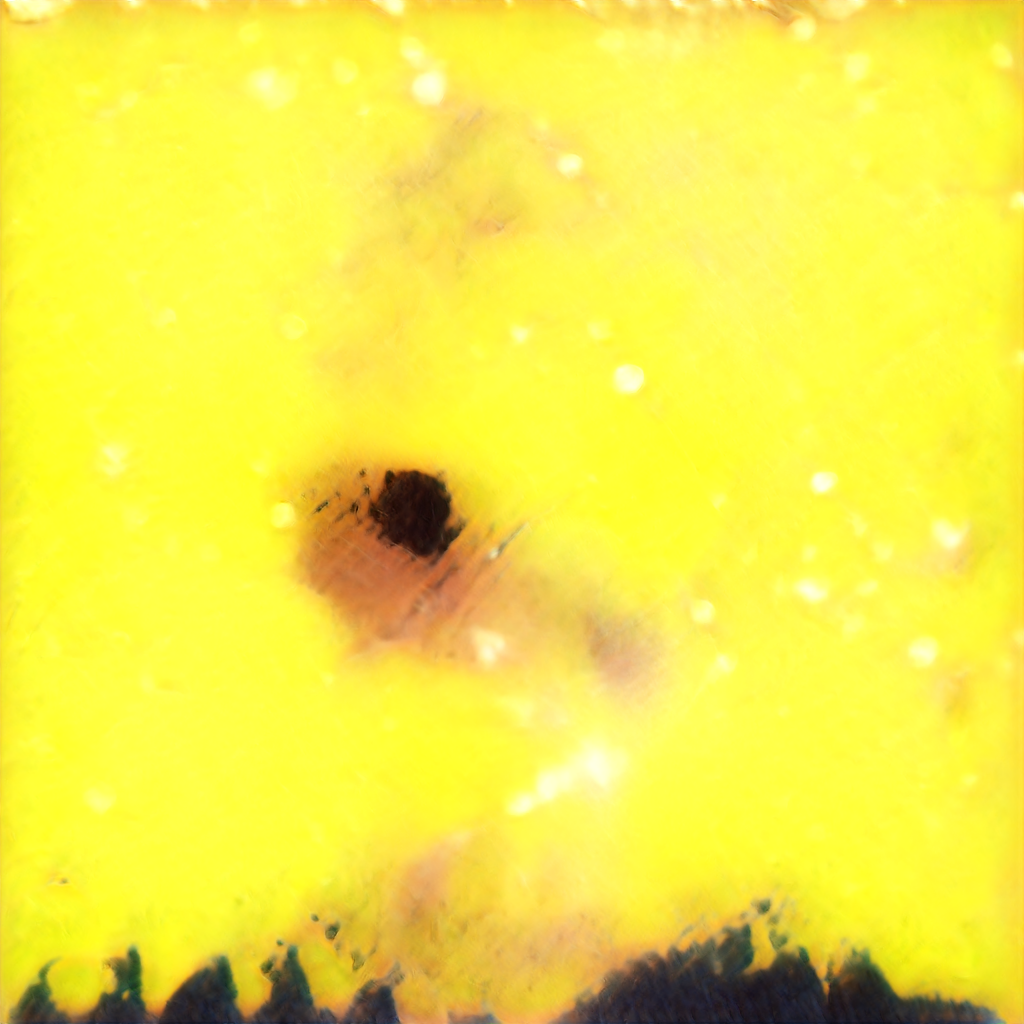}}
   \subfigure{\includegraphics[width=.25\textwidth]{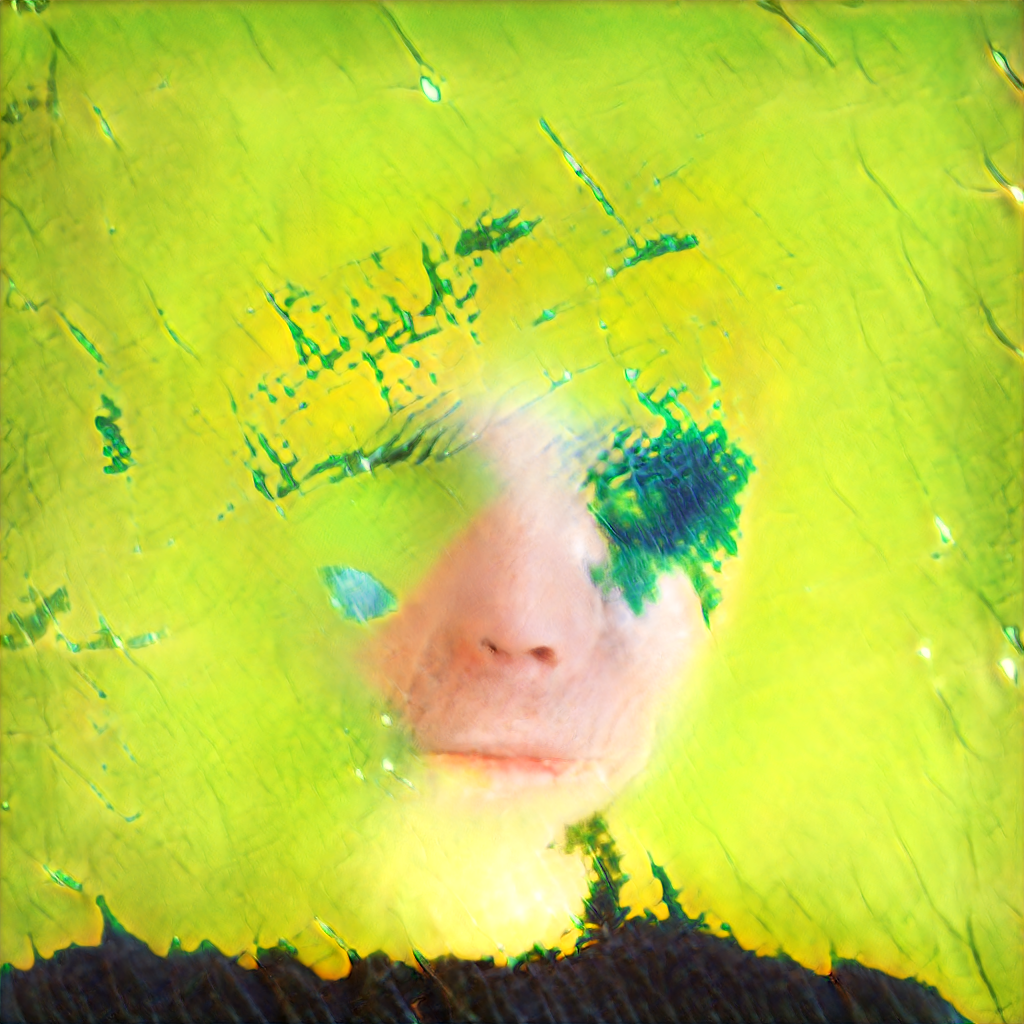}}
   \subfigure{\includegraphics[width=.25\textwidth]{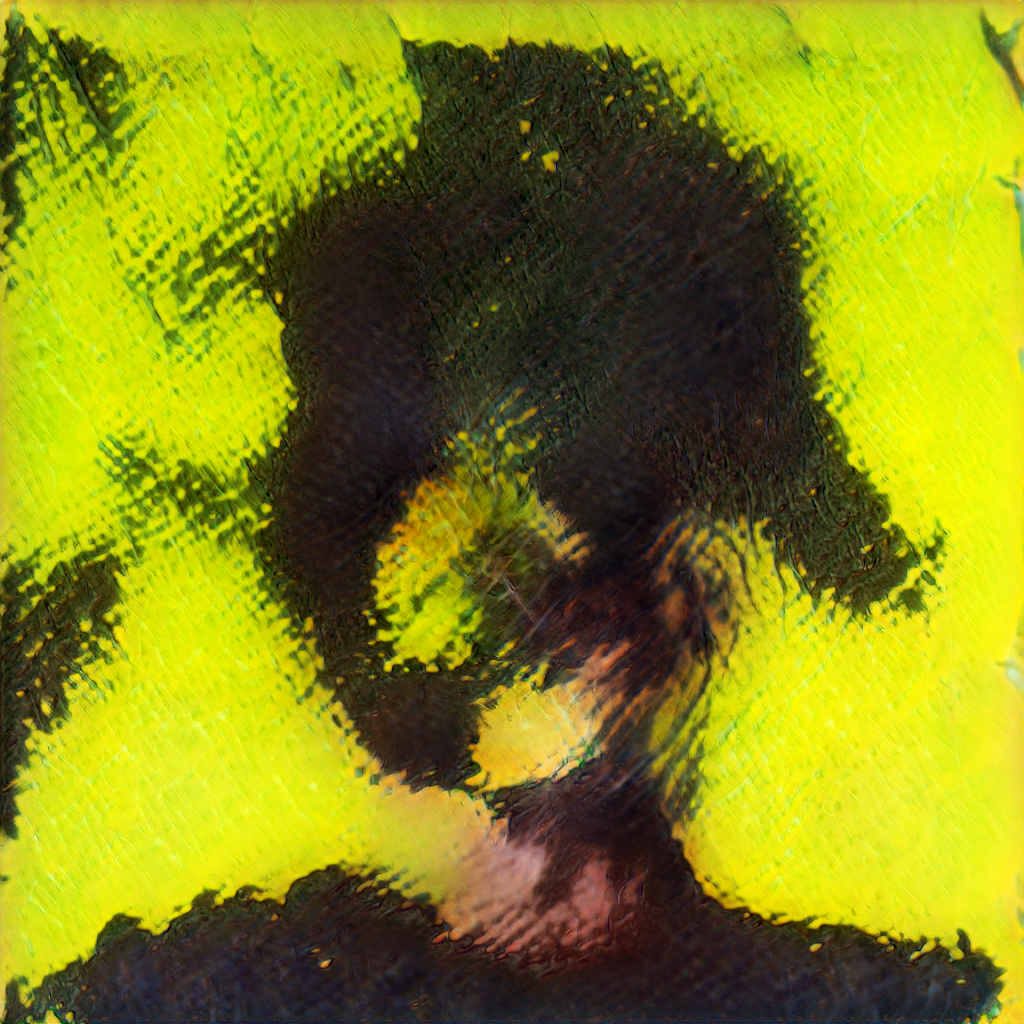}}
   \caption{Top left: the projected image into the FFHQ StyleGAN2 latent space of one of the band members of \emph{0171}. Other images: The 5 EP selected artworks that were made using the same latent code, but with multiple stochastic network bending transformations.%
   \textcopyright \emph{0171}, 2020.}
   \label{fig:stochastic_layers}
\end{figure}

A photograph of one of the band members was taken, and then the digital image was projected into the StyleGAN2 latent space of the model trained on the FFHQ dataset \cite{abdal2019image2stylegan,karras2019analyzing}. Different configurations of random layers were applied, and large batches of results were generated. When a configuration produced stylistically interesting and varied results, a hand-picked selection was saved, to later be shown to the band. Through a process of iteration from different input photographs, and different transform configurations, a final configuration was found that matched the aesthetic that the band wanted to convey.
After generating a large number of samples from this configuration, the best ones were highlighted and the band finally picked their favourite 5 samples, which were then used as the artworks for the singles in the EP (Figure \ref{fig:stochastic_layers}).

\section{Discussion}



\begin{figure}[tbp]
\noindent \begin{centering}
   \includegraphics[width=.2\textwidth]{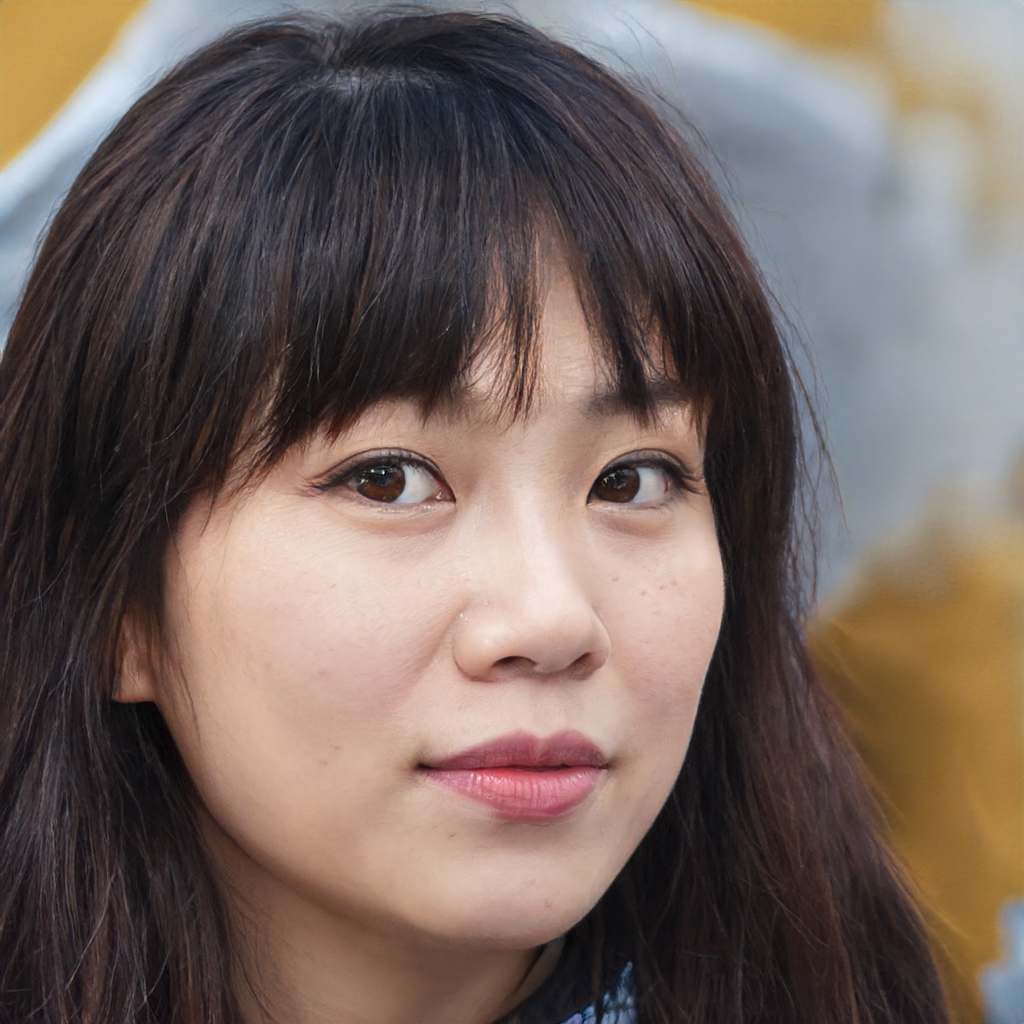}
   \includegraphics[width=.2\textwidth]{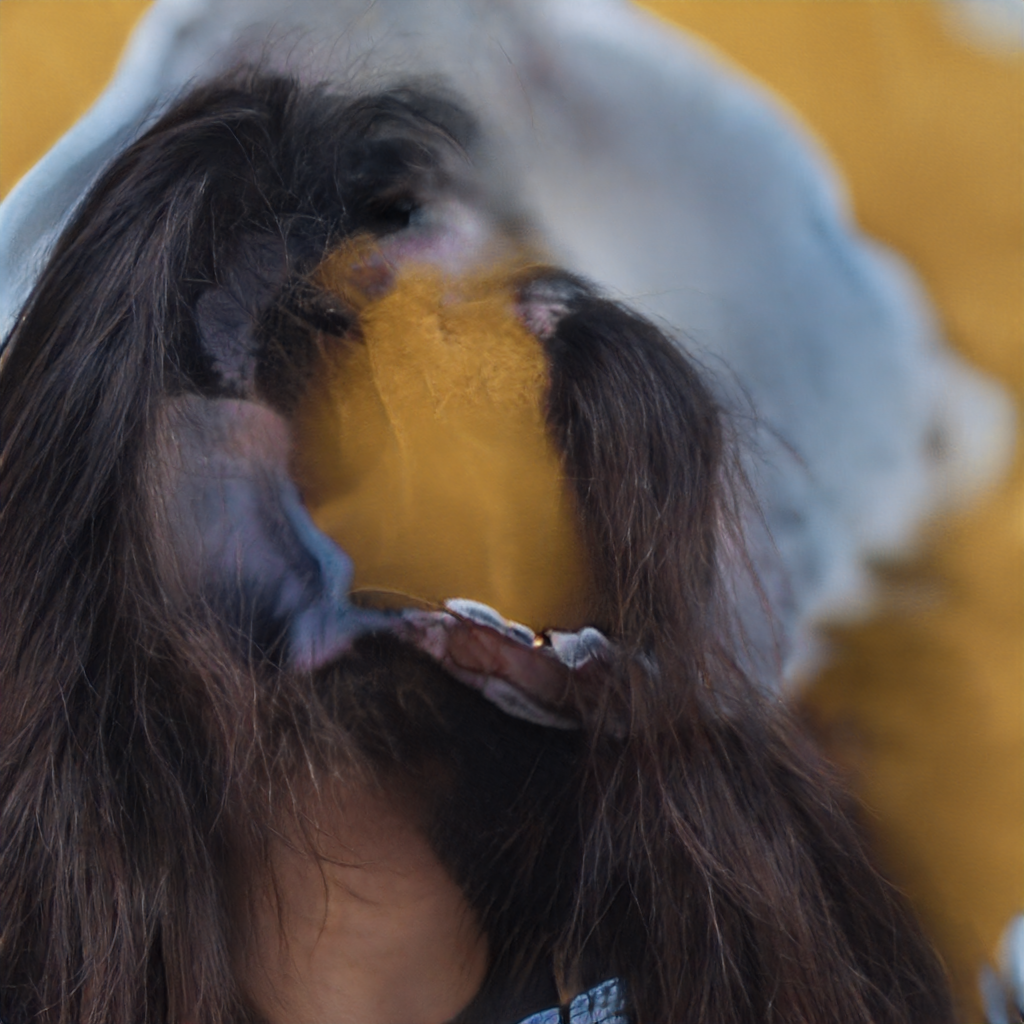}
   \includegraphics[width=.2\textwidth]{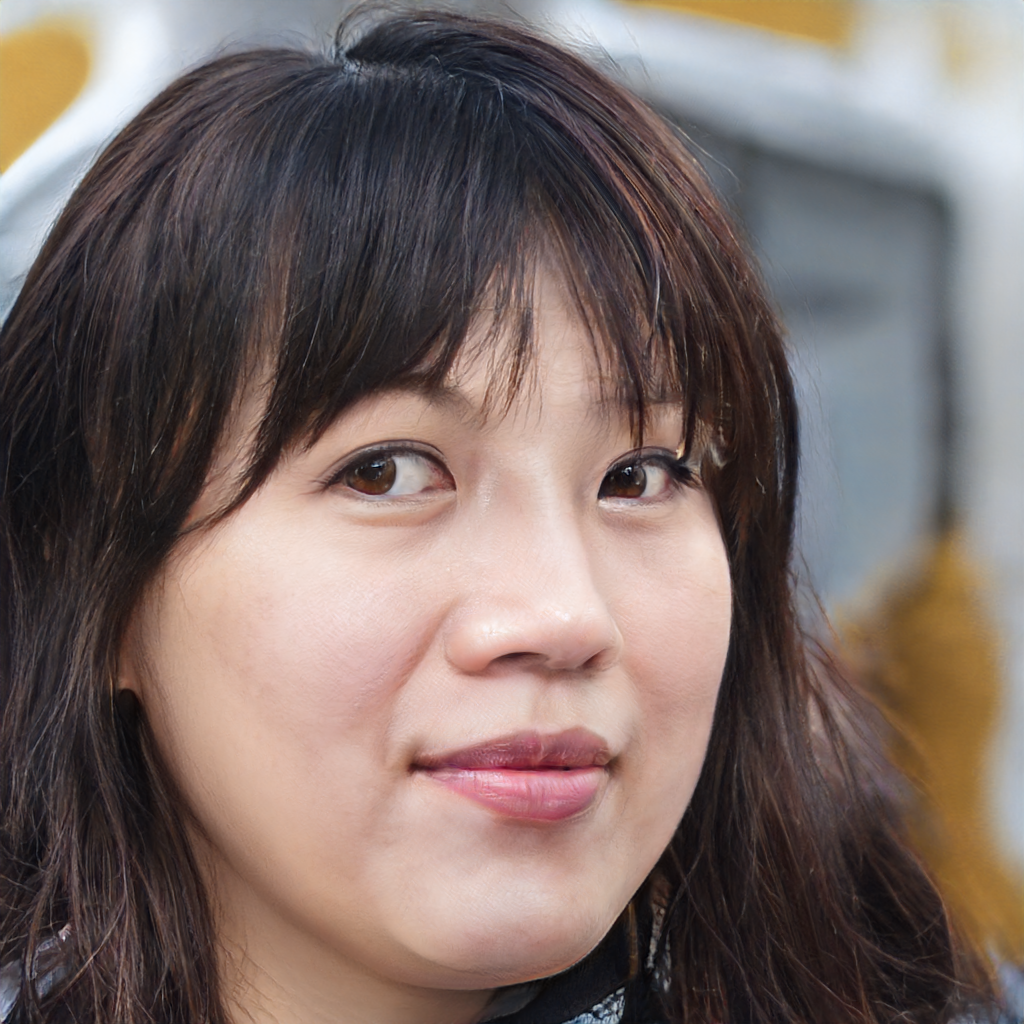}
   \includegraphics[width=.2\textwidth]{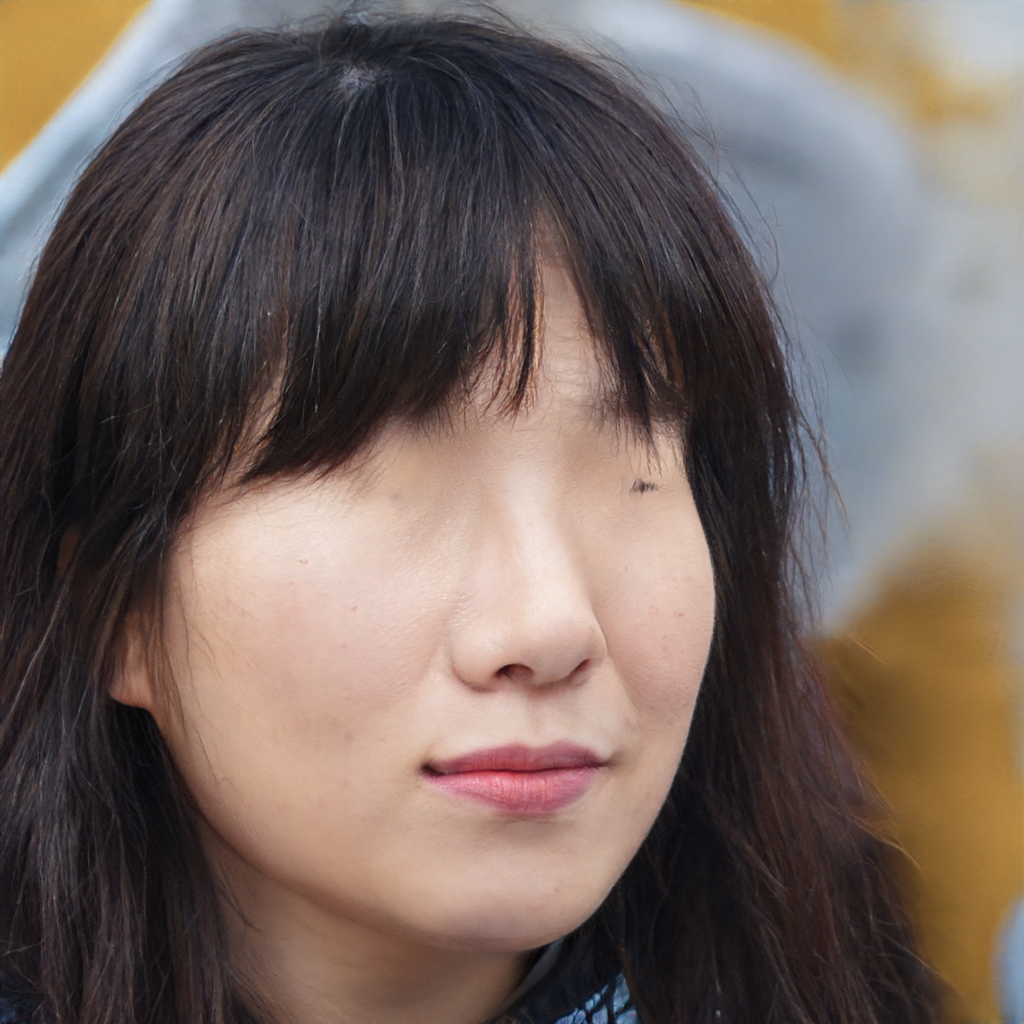}
   \par\end{centering}
   \noindent \begin{centering}
   \hfill (a)\hfill (b) \hfill (c) \hfill (d) \hfill\ %
   \par\end{centering}
   \smallskip
\noindent \begin{centering}
   \includegraphics[width=.2\textwidth]{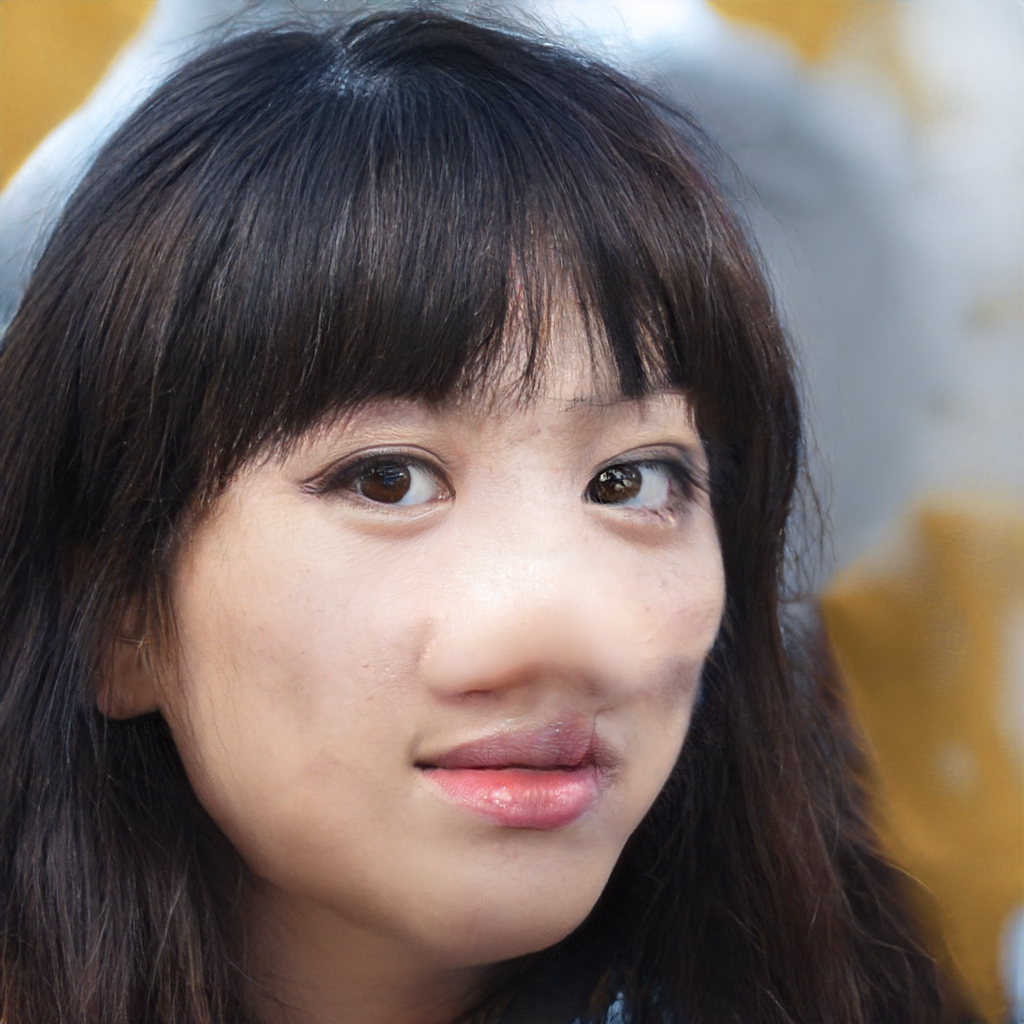}
   \includegraphics[width=.2\textwidth]{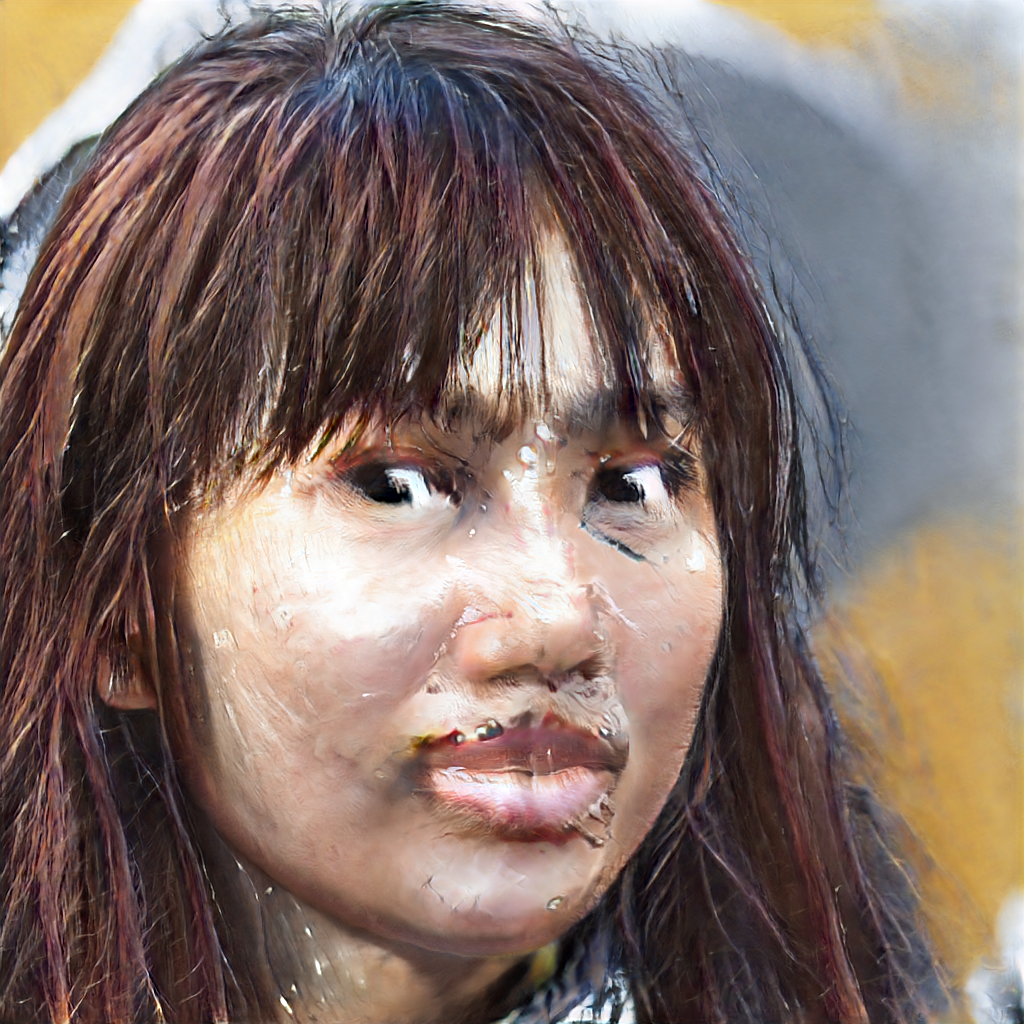}
   \includegraphics[width=.2\textwidth]{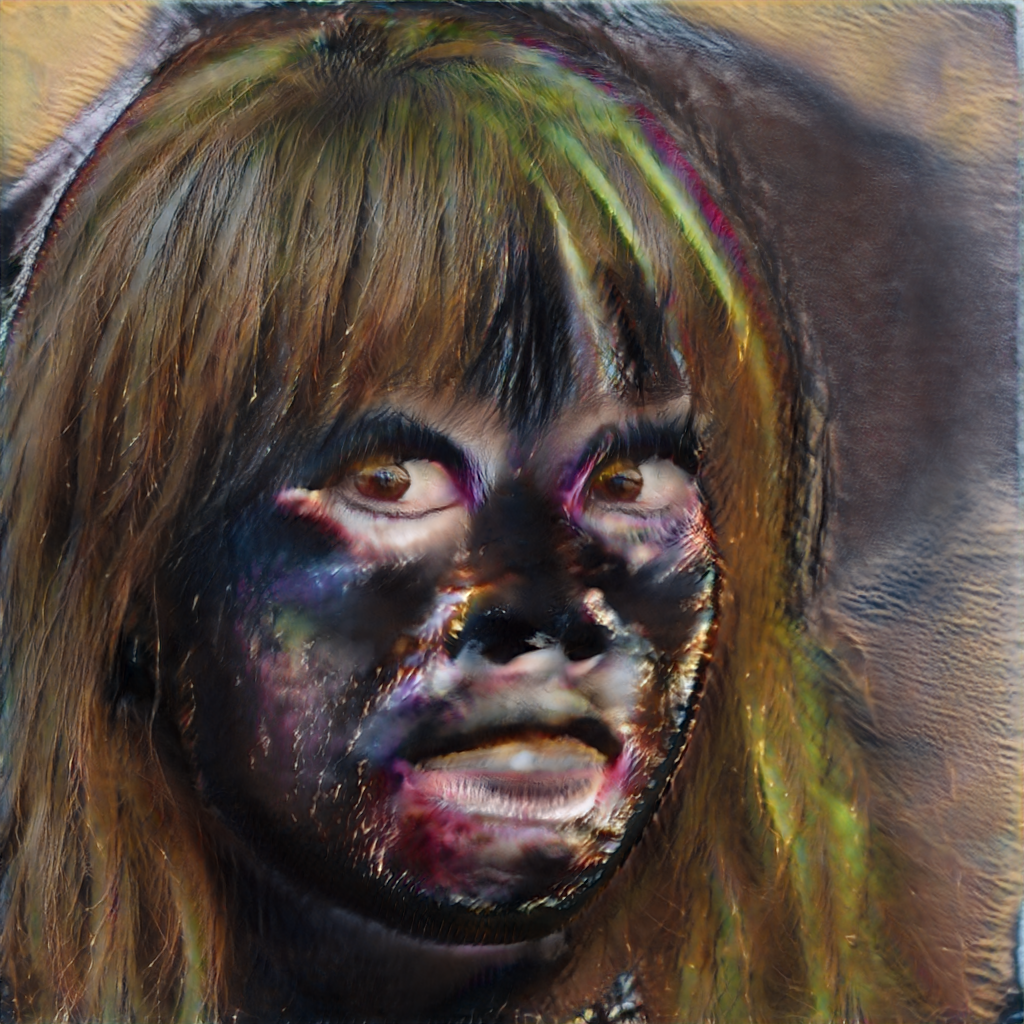}
   \includegraphics[width=.2\textwidth]{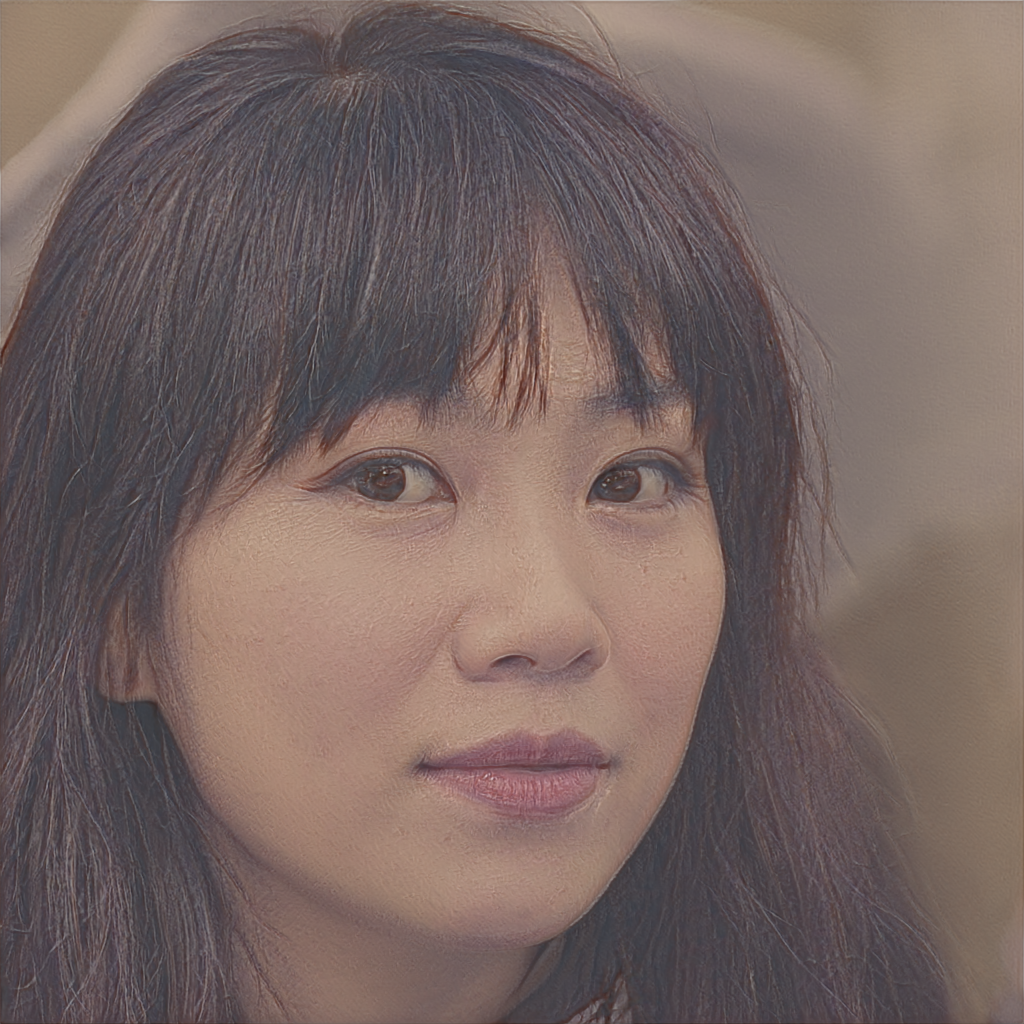}
   \par\end{centering}
   \noindent \begin{centering}
   \hfill (e)\hfill (f) \hfill (g) \hfill (h) \hfill\ %
   \par\end{centering}
   \caption{Clusters  of features in different layers of the model are responsible for the formation of different image attributes. (a) The unmanipulated result. (b) A cluster in layer 1 has been multiplied by a factor of -1 to completely remove the facial features. (c) A cluster in layer 3 has been multiplied by a factor of 5 to deform the spatial formation of the face. (d) A cluster in layer 6 has been ablated to remove the eyes. (e) A cluster in layer 6 has been dilated to enlarge the nose. (f) A cluster in layer 9 has been multiplied by a factor of 5 to distort the formation of textures and edges. (g) A cluster of features in layer 10 have been multiplied by a factor of -1 to invert the highlights on facial regions. (h) A cluster of features in layer 15 has been multiplied by a factor of 0.1 to desaturate the image. All transformations have been applied to sets of features discovered using our feature clustering algorithm (\S\ref{section:clustering}) in the StyleGAN2 model trained on the FFHQ dataset.}
   \label{fig:cluster_layer_comp}
\end{figure}

The main motivation of the clustering algorithm presented in this paper was to simplify the parameter space in a way that allows for more meaningful and controllable manipulations whilst also enhancing the expressive possibilities afforded by interacting with the system. Our results show that the clustering algorithm is capable of discovering groups of features that correspond to the generation of different semantic aspects of the results, which can then be manipulated in tandem. These semantic properties are discovered in an unsupervised fashion, and are discovered across the entire hierarchy of features present in the generative model. For example, Figure \ref{fig:cluster_layer_comp} shows the manipulation of groups of features across a broad range of layers that control the generation of: the entire face, the spatial formation of facial features, the eyes, the nose, textures, facial highlights and overall image contrast.

Grouping and manipulating features in a semantically meaningful fashion is an important component for allowing expressive manipulation. However, artists are often also ready to consider surprising, unexpected results, to allow for the creation of new aesthetic styles, which can become uniquely associated to an individual or group of creators. Therefore the tool needs to allow for unpredictable as well as predictable possibilities, which can be used in an exploratory fashion and can be mastered through dedicated and prolonged use \cite{dobrian2006nime}. There is usually a balance between utility and expressiveness of a system \cite{jacobs2017supporting}. While it will be required to build an interface and perform user studies to more conclusively state that our approach has struck such a balance, our current results do show that both predictable semantic manipulation and more unpredictable, expressive outcomes are possible. This is a good indication that our approach represents a good initial step, and with further refinements it can become an innovative powerful tool for producing expressive outcomes, when using deep generative models.

\subsection{Active Divergence}

One of the key motivations of our network bending approach, was to allow for the direct manipulation of generative models, such that the results were novel and divergent from the original training data, a goal that has been referred to as \textit{active divergence}  \cite{berns2020bridging}. One common criticism of using deep generative models in an artistic and creative context, is that they can only re-produce samples that \emph{fit} the distribution of samples in the training set. However, by introducing deterministic controlled filters into the computation graph during inference, these models can be used to produce a large array of novel results. Figure \ref{fig:layerwide_comparison} shows how the results vary drastically by applying the same transformation with the same parameters to different layers. Because our method alters the computational graph of the model, these changes to the results take effect across the entire distribution of possible results that can be generated. The results we have obtained markedly lie outside the distribution of training images, and allow for a very large range of possible outcomes, of which a small set of examples is seen in Figure \ref{fig:chaining_transforms}, which shows the broad range of outcomes possible when various transformation applied to different sets of features in different layers are combined. We emphasise that such outcomes, to the best of our knowledge, could not reasonably be produced using any other existing method of image manipulation or generation.

\subsection{Comparison with Other Methods}

With respect to the semantic analysis and manipulation of a generative model, our approach of clustering features and using a broad array of transformation layers is a significant advance over previous works \cite{Bau2017-vg,Bau2018-td,bau2019semantic, Brink2019-gc}. This recent thread of techniques only interrogate the function of individual features, and as such are unlikely to be capable of capturing a full account of how a deep network generates results, since such networks tend to be robust to the transformation of individual features. 

\begin{figure}[!ht]
   \centering
   \subfigure{\includegraphics[width=.25\textwidth]{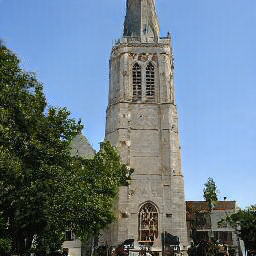}}
   \subfigure{\includegraphics[width=.25\textwidth]{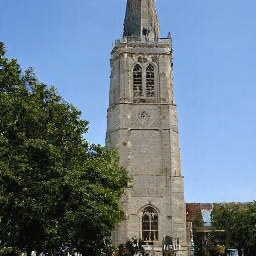}}
   \subfigure{\includegraphics[width=.25\textwidth]{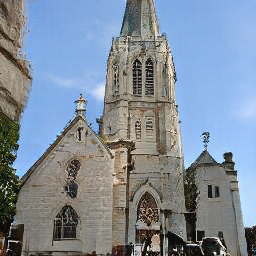}}
   \caption{Groups of features that are not particularly sensitive to ablation may be more sensitive to other kinds of transformation. Left: original unmodified input. Middle: a cluster of features in layer 3 that has been ablated. Right: the same cluster of features that has been multiplied by a scalar of 5. As can be seen ablation had a negligible effect, only removing a small roof structure which was behind the foliage. On the other hand, multiplying by a factor of 5 removes the trees whilst altering the building structure to have gable roof sections on both the left and right sides of the church - which are now more prominent and take precedence in the generative process. Samples are taken from the StyleGAN2 model trained on the LSUN church dataset.}
   \label{fig:ablation_comp}
\end{figure}

We also show that sets of features, which may not be particularly responsive to certain transformations, are very responsive to others. Figure \ref{fig:ablation_comp} shows that in the model trained on the LSUN church dataset, a cluster of features, that when ablated has little noticeable effect on the result, can produce significant changes when using another transformation on the same cluster, here removing the trees and revealing the church building that was obscured by the foliage in the original result. This, we argue, shows that the functionality of features, or sets of features, cannot be understood only through ablation (which is the approach used in GAN dissection \cite{Bau2018-td}), because of the high levels of redundancy present in the learned network parameters. We show that their functionality can be better understood by applying a wide range of deterministic transformations, of which different transformations are better suited to revealing the utility of different sets of features (Figures \ref{fig:cluster_layer_comp} \& \ref{fig:ablation_comp}).

Our method of analysis is completely \emph{unsupervised}, and does not rely on auxiliary models trained on large labelled datasets (such as in \cite{Bau2018-td, isola2017image, park2019semantic}) or other kinds of domain specific knowledge. This approach therefore can be applied to any CNN based GAN architecture used for image generation which has been trained on any dataset. This is of particular relevance to artist who create their own datasets and would want to apply these techniques to models they have trained on their own data. Labelled datasets, especially the pixel labelled datasets used in semantic image synthesis, are prohibitively time consuming (or expensive) to produce for all but a few artists or organisations. Having a method of analysis that is completely unsupervised and can be applied to unconditional generative models is important in opening up the possibility that such techniques become adopted more broadly.

The framework we have presented is the first approach to manipulating generative models that focuses on allowing for a large array of novel expressive outcomes. In contrast to other methods that manipulate deep generative models \cite{bau2019semantic, bau2020rewriting}, our approach allows the manipulation of any feature or set of features in any layer, with a much broader array of potential transformations. By allowing for the combination of many different transformations, it is evident that the outcomes can diverge significantly from the original training data, allowing for a much broader range of expressive outcomes and new aesthetic styles than would be possible with methods derived from semantic image synthesis \cite{isola2017image, chen2017photographic, park2019semantic} or semantic latent manipulation \cite{brock2016neural,  shen2020interpreting, harkonen2020ganspace}.
    


\section{Conclusion and Future Work}

In this paper we have introduced a novel approach for the interaction with and manipulation of deep generative models that we call \textit{network bending} which we have demonstrated on generative models trained on several datasets. By inserting deterministic filters inside the network, we present a framework for performing manipulation inside the networks' black-box and utilise it to generate samples that have no resemblance to the training data, or anything that could be created easily using conventional media editing software. We also present a novel clustering algorithm that is able to group sets of features, in an unsupervised fashion, based on spatial similarity of their activation maps. We demonstrated that this method is capable of finding sets of features that correspond to the generation of a broad array of semantically significant aspects of the generated images. This provides a more manageable number of sets of features that a user could interact with. We propose that using our approach, possibly in conjunction with other methods, for a better understanding and navigating of the latent space of a model, can provide a very powerful set of tools for the production of novel and expressive images. 


In future work we plan to build an interface around this framework, and to perform user studies to understand how artists would want to and how they end up using this framework, and to refine the parameter space to allow for a balance between utility and expressiveness of possible outcomes. At the time of publication, this network bending framework ---inserting deterministically controlled transformation layers and applying them to clustered sets of features--- has been adapted and applied in the domains of audio synthesis \cite{mccallum2020network} and audio-reactive visual synthesis \cite{brouwer2020audio}. In future work we look to further extend this framework to generative models of other domains such as those that produce text, video or 3D images and meshes.

%

\subsubsection*{Acknowledgements}
This work has been supported by UK’s EPSRC Centre for Doctoral Training in Intelligent Games
and Game Intelligence (IGGI; grant EP/L015846/1).

\bibliographystyle{unsrt}
\bibliography{main}

\end{document}